# Haploid-Diploid Evolutionary Algorithms:
# The Baldwin Effect and Recombination Nature's Way

**Larry Bull**[1]

**Abstract.** This paper uses the recent idea that the fundamental haploid-diploid lifecycle of eukaryotic organisms implements a rudimentary form of learning within evolution. A general approach for evolutionary computation is here derived that differs from all previous known work using diploid representations. The primary role of recombination is also changed from that previously considered in both natural and artificial evolution under the new view. Using well-known abstract tuneable models it is shown that varying fitness landscape ruggedness varies the benefit of the new approach.

## 1 INTRODUCTION

The vast majority of work within evolutionary computation has used a haploid representation scheme; individuals are one solution to the given problem. Simple prokaryotic organisms similarly contain one set of genes, whereas the more complex eukaryotic organisms – such as plants and animals - are predominantly diploid and contain two sets of genes. A small body of work exists using a diploid representation scheme within evolutionary computation; individuals carry two solutions to the given problem. In all but one known example, a dominance scheme is utilized to reduce the diploid down to a traditional haploid solution for evaluation. That is, as individuals carry two sets of genes, a heuristic is included to choose which of the genes to use (see [1] for a recent review). The only known exception is work by Hillis [2] on sorting networks wherein non-identical solution values at a given gene position are both used to form the solution for evaluation, thereby enabling solution lengths to vary up to the combined length of the two constituent genomes.

However, in nature, eukaryotes exploit a haploid-diploid cycle where haploid cells form a diploid cell/organism. At the point of reproduction by the cell/organism, the haploid genomes within the diploid each form haploid gamete cells that (may) join with a haploid gamete from another cell/organism to form a diploid (Figure 1). Specifically, each of the two genomes in an organism is replicated, with one copy of each genome being crossed over. In this way copies of the original pair of genomes may be passed on, mutations aside, along with two versions containing a mixture of genes from each. Previous explanations for the emergence of the alternation between the haploid and diploid states are typically based upon its being driven by changes in the environment (after [3]). Recently, an explanation for the haploid-diploid cycle in eukaryotes has been presented [4] which also explained other aspects of their sexual reproduction, including the use of recombination, based upon the Baldwin effect [5].

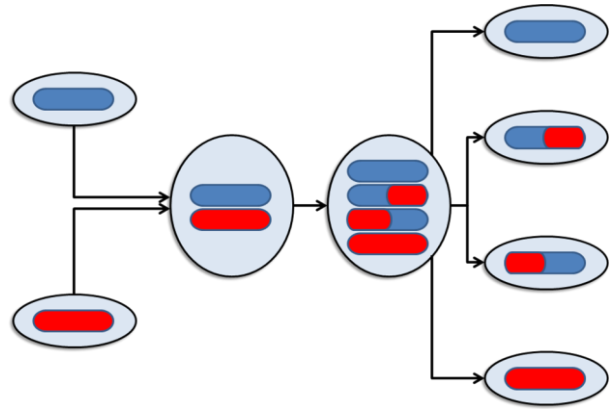

**Figure 1.** The haploid-diploid phases within the two-stage meiosis process seen in most eukaryotic organisms.

The Baldwin effect is the existence of phenotypic plasticity which enables an organism to display a different (better) fitness than its genome directly represents. Importantly, such learning can affect (improve) the evolutionary process by altering the shape of the underlying fitness landscape. For example, if a very poor combination of genes is able to consistently learn a far superior phenotype, their frequency will increase under selection more than expected without learning; the effective shape of the fitness landscape will be changed for the poor gene combination. Becoming diploid can potentially alter gene expression in comparison to being haploid and hence affect the expected phenotype of each haploid alone since *both genomes are active in the cell* - through changes in gene product concentrations, partial or co-dominance, etc. That is, *the fitness of the diploid cell/organism is a combination of the fitness contributions of the composite haploid genomes*. If the cell/organism subsequently remains diploid and reproduces asexually, there is no scope for a rudimentary Baldwin effect. However, if there is a reversion to a haploid state for reproduction, there is the potential for a significant mismatch between the utility of the haploids passed on compared to that of the diploid selected; *individual haploid gametes do not contain all of the genetic material through which their fitness was determined*. That is, the effects of haploid genome combination into a diploid can be seen as a simple form of phenotypic plasticity for the individual haploid genomes before they revert to a solitary state during reproduction.

In the haploid case, variation operators generate a new genome at a *point* in the fitness landscape. Under the haploid-diploid cycle, two such genomes join and their fitness as a diploid is a function of their individual genomes and any interactions. In the simplest case, the fitness of the diploid can be taken as the average fitness of the individual haploids. Hence the fitness of the diploid is a point between the fitness of the individual haploids. Since the constituent haploid genomes are each assigned that fitness under selection, the shape of the underlying haploid genome fitness landscape can be changed. That is, *evolution forms a generalization about the typical fitness of solutions found between the two haploid genomes*.

---

[1] Dept. of Computer Science & Creative Technologies, Univ. of the West of England, BS16 1QY, UK. Email: larry.bull@uwe.ac.uk.

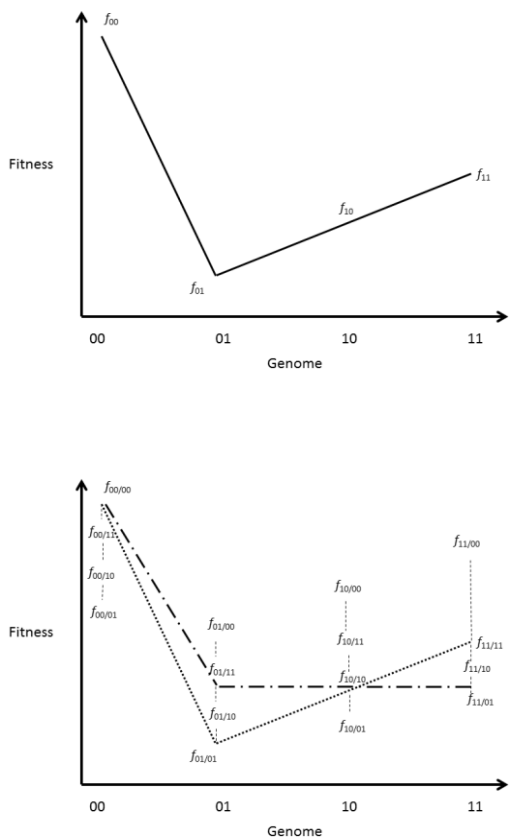

**Figure 2.** Comparing the fitnesses of haploid genomes under the haploid-diploid case (bottom) to the traditional haploid case (top). The case where 01-11 are always paired and the others pair homogeneously is shown.

Figure 2 shows a very simple example of the contrast between the standard haploid genome landscape evolution view and how the haploid-diploid cycle alters the landscape with fitness contribution averaging. Under the haploid-diploid cycle the apparent fitness of the valley is potentially increased, increasing the likelihood selection will maintain such haploid genomes within the population, thereby increasing the likelihood of the valley being crossed to the optimum. This suggests an increased benefit from the haploid-diploid cycle as landscape ruggedness increases. It can also be noted that the shape of the fitness landscape *varies* based upon the haploid genomes which exist within a given population at any time. This is also significant since, as has been pointed out for coevolutionary fitness landscapes [6], such movement potentially enables the *temporary* creation of neutral paths, where the benefits of (static) landscape neutrality are well-established (after [7]).

Moreover, the variation operators can then be seen to change the *bounds* for sampling combined genomes within the diploid fitness landscape by altering the distance between the two end points in the underlying haploid fitness landscape. That is, the degree of possible change in the distance between the two haploid genomes controls the amount of learning possible per cycle. It has long been known that the most beneficial amount of learning under the Baldwin effect increases with the ruggedness of the fitness landscape [8]. Numerous explanations exist for the benefits of recombination in both natural (eg, [9]) and artificial systems (eg, [10]), the latter focusing solely upon haploid genomes and neither considering the potential Baldwin effect under the haploid-diploid cycle. The effects of recombination in the haploid-diploid cycle become clear with the new insight: *recombination potentially moves the current end points significantly, thereby increasing the rate of change in the fitness level generalizations in comparison to that possible under typical gene value mutation rates alone*. This paper presents a new class of evolutionary algorithm which exploits the new understanding of the haploid-diploid cycle in eukaryotes.

## 2 NK AND RBNK MODELS

The NK model of fitness landscapes to allow the systematic study of various aspects of evolution (see [11]). In the model an individual is represented by a set of $N$ (binary) genes or traits, each of which depends upon its own value and that of $K$ randomly chosen others in the individual. Thus increasing $K$, with respect to $N$, increases the epistatic linkage. This increases the ruggedness of the fitness landscapes by increasing the number of fitness peaks. The model assumes all epistatic interactions are so complex that it is only appropriate to assign (uniform) random values to their effects on fitness. Therefore for each of the possible $K$ interactions, a table of $2^{(K+1)}$ fitnesses is created, with all entries in the range 0.0 to 1.0, such that there is one fitness value for each combination of traits. The fitness contribution of each trait is found from its individual table. These fitnesses are then summed and normalised by $N$ to give the selective fitness of the individual (Figure 3).

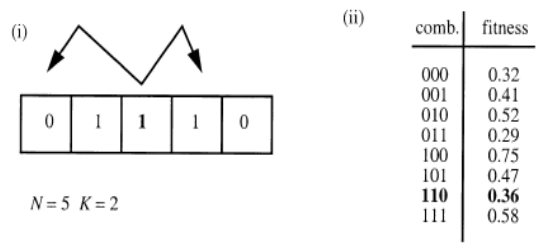

**Figure 3.** An example NK model. (i) shows how the fitness contribution of each gene depends on $K$ random genes. Therefore there are $2^{(K+1)}$ possible allele combinations, each of which is assigned a random fitness as shown in (ii). Each gene of the genome has such a table created for it. Total fitness of a given genome is the normalized sum of these values.

In the RBN model of genetic regulatory networks (see [11]) a network of $R$ nodes, each with $B$ directed connections from other nodes in the network all update synchronously based upon the current state of those $B$ nodes. Hence those $B$ nodes are seen to have a regulatory effect upon the given node, specified by the given Boolean function attributed to it. Nodes can also be self-connected. Since they have a finite number of possible states and they are deterministic, such networks eventually fall into an attractor. It is well-established that the value of $B$ affects the emergent behaviour of RBN wherein attractors typically contain an increasing number of states with increasing $B$. Three phases of behaviour exist: ordered when $B=1$, with attractors consisting of one or a few states; chaotic when $B>2$, with a very large number of states per attractor; and, a critical regime at $B=2$, where similar states lie on trajectories that tend to neither diverge nor converge (see [11] for discussions of this critical regime, e.g., with respect to perturbations).

The combination of the RBN and NK models enables the exploration of the relationship between phenotypic traits and the genetic regulatory network by which they are produced – the RBNK model [12]. In this paper, the following simple scheme is adopted: $N$ phenotypic traits are attributed to randomly chosen nodes within the network of $R$ genes (Figure 4). Thereafter all aspects of the two models remain as described, with simulated evolution used to evolve the RBN on NK landscapes. Hence the NK element creates a tuneable component to the overall RBN fitness landscape.

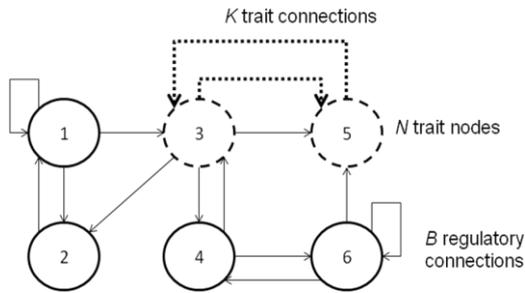

An $R=6$, $B=2$, $N=2$, $K=1$ network

**Figure 4**. Example RBNK model. Each network consists of $R$ nodes, each node containing $B$ integers in the range $[1, R]$ to indicate input connections and a binary string of length $2^B$ to indicate the Boolean logic function over those connections.

## 3 A HAPLOID-DIPLOID ALGORITHM

In the simple haploid-diploid evolutionary algorithm (HD-EA) used here, a population member is diploid and so contains two solutions to the given task. The fitness of a population member is the average of the fitness of its two genomes/solutions. Two parents are selected to form one offspring here, following the process shown in Figure 1: one parent is selected and copies each of its haploid genomes; one copy of each of the two genomes is crossed over to create two new variants; mutation is applied to all four genomes; and finally, one of the four resulting haploid genomes is picked at random to form half of the new offspring individual. The process is repeated with the second parent to form the other half of the offspring:

```
BEGIN
INITIALISE population each with two random solutions
EVALUATE each candidate's pair of solutions
REPEAT UNTIL (TERMINATION CONDITION) DO
        FOR each new candidate solution DO
            REPEAT twice DO
                SELECT parent candidate
                COPY candidate with ERROR
                COPY candidate copy again
                RECOMBINE two non-sister solutions
                CHOOSE one of the four solutions
            OD
            EVALUATE new solution
            REPLACE IF (UPDATE CONDITION)
                candidate with new solution
        OD
OD
END
```

The details of the selection process, recombination and mutation operators are not specific to the general approach. Specific examples are next described for the chosen tasks.

## 4 EXPERIMENTATION

A standard steady-state evolutionary process is used here: reproduction selection is via a binary tournament (size 2), replacement selection uses the worst individual, the population contains 50 diploid individuals, the constituent haploid genomes are binary strings of length $N=50$, mutation is deterministic at rate $1/N$ and single-point crossover is used. Since each diploid individual requires two evaluations, the comparative traditional haploid scheme is run for twice as many generations, with all other details the same. The results presented are the average fitness of the best individual after 20,000 generations (40,000 for the haploid), for various $K$. Each experiment consists of running ten random populations on each of ten fitness landscapes, ie, results are the average of 100 runs. As can be seen in Figure 5, for $K>4$ the HD-EA is better than the traditional haploid approach (T-test, $p<0.05$), otherwise performance is equal (T-test, $p\geq0.05$).

Most details of the experimentation remain unchanged from the NK model for the RBNK model. Individual solutions are now integer strings of length ($R=100$) and mutation is applied deterministically at a rate $1/R$. In contrast to bit-flipping, mutation can either alter the Boolean function of the randomly chosen node or alter a randomly chosen connection for that node (equal probability). Following the known behaviour of $B=2$ RBN discussed above, only $B=2$ is used here. A fitness evaluation is ascertained by first assigning each node to a randomly chosen start state and updating each node synchronously for $T$ cycles ($T=50$). Here $T$ is chosen such that the networks have typically reached an attractor. At update cycle $T$, the value of each of the $N$ trait nodes is then used to calculate fitness on the given NK landscape. This process is repeated ten times on the given NK landscape iteration. Each experiment again consists of running ten random populations on each of ten fitness landscapes, ie, results are here the average of 1000 runs (after [12]). Figure 6 shows for $K \geq 6$ the HD-EA is better than the traditional haploid approach (T-test, $p<0.05$), otherwise performance is equal (T-test, $p \geq 0.05$).

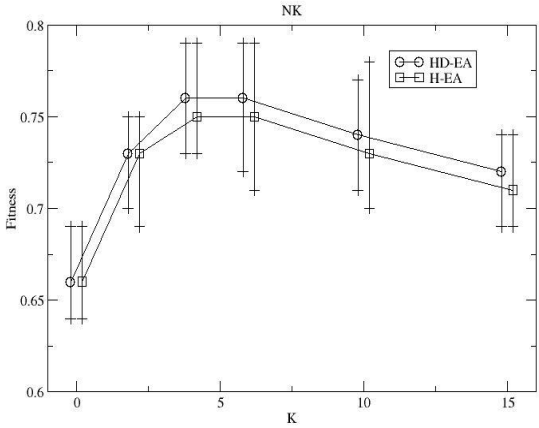

**Figure 5**. Comparative performance of the best individual evolved using the haploid-diploid algorithm (HD-EA) to the traditional haploid approach (H-EA), over a range of fitness landscapes ($N$=50). Error bars show max and min values.

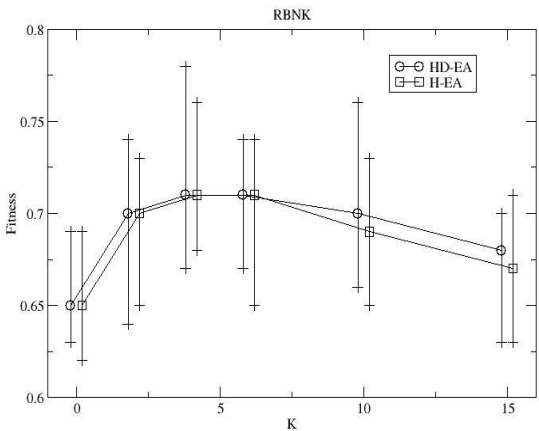

**Figure 6**. Comparative performance of the best RBN individual evolved using the haploid-diploid algorithm (HD-EA) to the traditional haploid approach (H-EA), over a range of fitness landscapes ($R$=100, $B$=2, $N$=50). Error bars show max and min values.

## 6 CONCLUSIONS & FUTURE WORK

This paper has presented a new form of evolutionary algorithm which exploits the haploid-diploid cycle seen in eukaryotic organisms. In contrast to previous work on diploid representations, both candidate solutions are evaluated and a combined fitness attributed to the individual. Since the diploid is reduced to a haploid under reproduction, there is a mismatch between the actual utility of the haploid and the fitness attributed to it. This is seen as exploiting a rudimentary form of the Baldwin effect, with the diploid phase seen as the "learning" step [4]. The scheme has been shown beneficial as the ruggedness of the fitness landscape increases for both a traditional binary-encoded optimization task and a dynamical network design task.

Every generation was assumed sexual here, whereas some eukaryotes can also reproduce asexually. Future work should consider this fact by introducing a new parameter to introduce (traditional) haploid cycles. The effects of varying the number of copies of genomes at the point of reproduction should also be explored – the typical natural case of pre-meiotic doubling was used here. The use of many recombination schemes would also seem to hold the potential to improve performance, along with other search operators from the literature. Results (not shown) typically find improvement ($K$>0) for both the haploid and diploid algorithms using one-point crossover on the RBNK but not the NK tasks. Note the potential benefits of crossover as traditionally understood (eg, [10]) are not lost under the new scheme.